\documentclass{article}

\PassOptionsToPackage{numbers, compress}{natbib}

\usepackage[preprint]{neurips_2022}

\usepackage[utf8]{inputenc} 
\usepackage[T1]{fontenc}    
\usepackage{hyperref}       
\usepackage{url}            
\usepackage{booktabs}       
\usepackage{amsfonts}       
\usepackage{amsmath}
\usepackage{nicefrac}       
\usepackage{microtype}      
\usepackage{xcolor}         
\usepackage{threeparttable}

\usepackage[ruled,vlined]{algorithm2e}
\usepackage{caption}
\usepackage{subcaption}
\usepackage{graphicx}

\title{NN2Rules: Extracting Rule List from Neural Networks}

\author{%
  G Roshan Lal \\ 
  LinkedIn AI\\
  Mountain View, CA, US\\
  \texttt{rlal@linkedin.com} \\
  \And 
  Varun Mithal\\
  LinkedIn AI\\
  Mountain View, CA, US\\
  \texttt{vamithal@linkedin.com} \\
}

\newcommand{\cut}[1]{}

\begin{document}

\maketitle

\begin{abstract}
We present an algorithm, NN2Rules, to convert a trained neural network into a rule list. Rule lists are more interpretable since they align better with the way humans make decisions. NN2Rules is a decompositional approach to rule extraction, i.e., it extracts a set of decision rules from the parameters of the trained neural network model. We show that the decision rules extracted have the same prediction as the neural network on any input presented to it, and hence the same accuracy. A key contribution of NN2Rules is that it allows hidden neuron behavior to be either soft-binary (eg. sigmoid activation) or rectified linear (ReLU) as opposed to existing decompositional approaches that were developed with the assumption of soft-binary activation. 
\end{abstract}

\section{Introduction}

In recent years, many decision systems have been constructed as black box machine learning models such as Neural Networks. Lack of understanding of the internal logic of decision systems, especially those used for critical tasks, constitutes both a practical and an ethical issue. A risk is the possibility of making wrong
decisions, learned from artifacts or spurious correlations in the training data. In addition, the European Parliament adopted the General Data Protection Regulation (GDPR) in 2018 which introduced a right of explanation for all individuals to obtain ``meaningful explanations of the logic involved'' when automated decision making takes place. Due to the above reasons there has been a surge in the tools for  understanding AI models. 

The research for understanding how a black box works can be broadly placed into two categories: (i) the problem of designing a transparent model to solve the same problem with similar performance (Transparent Box Design), and  (ii) the problem of explaining how a black-box decision system works (Black Box Explanation). Moreover, the Black Box Explanation problem can be further divided among: (a) Outcome Explanation- where the goal is to understand the reasons for the decisions on a given instance, and (b) Model Explanation- where the goal is to provide a global explanation for the entire logic of the black-box model.

\begin{figure}[ht]
\centering
\includegraphics[width=0.67\columnwidth]{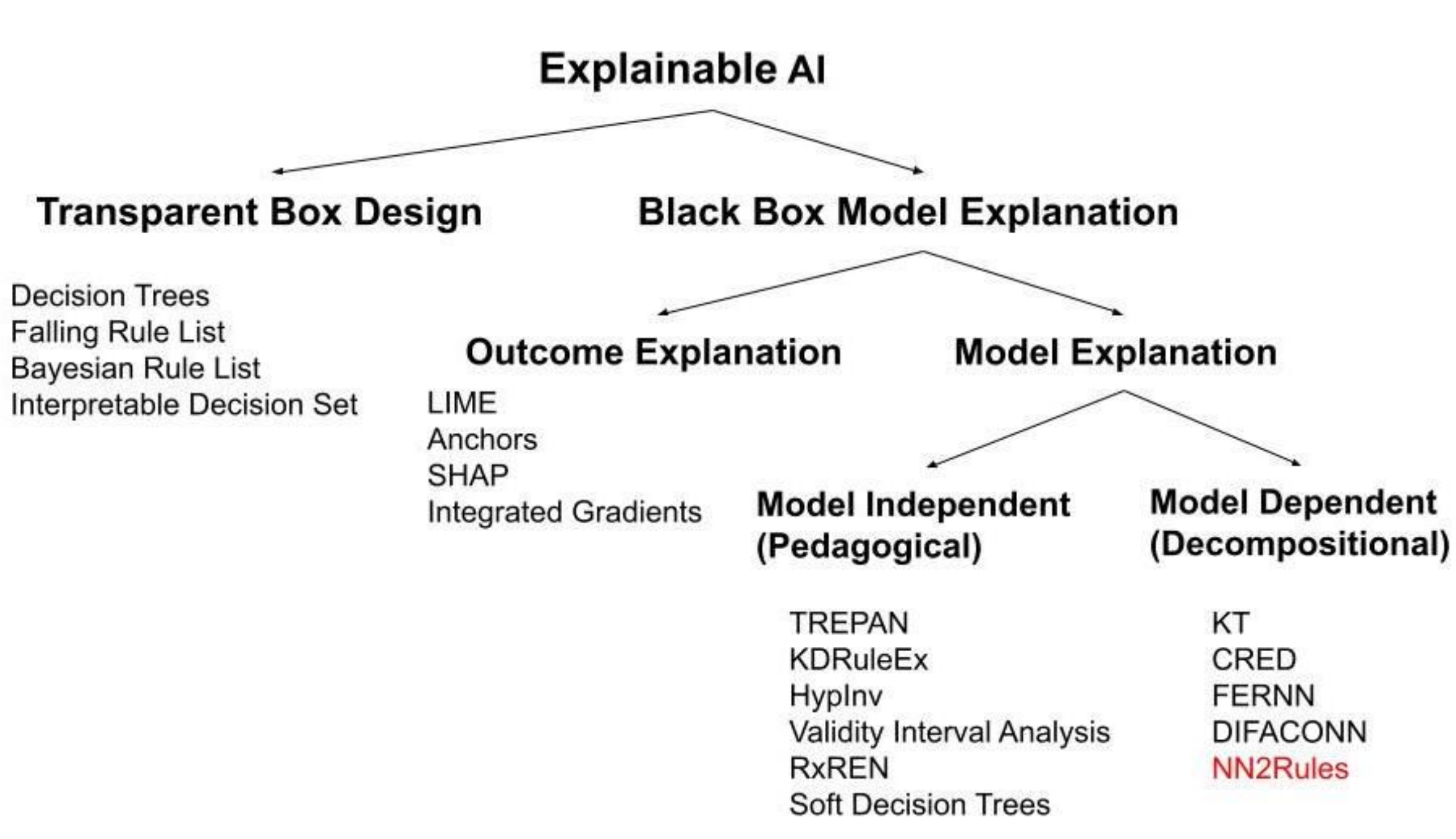} 
\vspace{-2mm}
\caption{A taxonomy of Explainable AI tasks}.
\label{fig1}
\vspace{-7mm}
\end{figure}

We focus on the \emph{Model Explanation} problem which aims to learn an interpretable model that mimics the behavior of the black box (\emph{fidelity}) and is understandable by humans (\emph{comprehensible}). Popular choice of interpretable models include decision trees, or decision lists. There are two broad categories of model explanation techniques for neural networks \cite{Andrews1995373}:
\begin{itemize}
    \item \emph{Decompositional/ Model dependent}: Techniques that explain the decisions made by individual neurons in a neural network and then put them together to explain the decisions made by the network architecture. They rely on specific architecture of the models (eg. choice of activation functions, etc.) hence often referred to as Model Dependent approaches. 
    \item \emph{Pedagogical/ Model independent}: Techniques that consider the neural network as a black box, and explain the input-output behavior of the black-box model as a whole. 
\end{itemize}

Decompositional approaches have higher fidelity but poorer comprehensibility compared to pedagogical approaches. \cite{huysmans2006using}. Therefore, often pedagogical approaches are used to obtain a summary of the model behavior, while decompositional approaches are used when the goal is to understand behavior of the neural network more comprehensively.

\textbf{Our Contribution:} In 2011 \cite{glorot2011deep} found Rectified Linear Unit (ReLU) enables better training of deeper networks, compared to the then widely used soft-binary activation functions (e.g., the logistic sigmoid and the hyperbolic tangent). As a consequence, ReLU is the most popular activation function for neural networks today. However, existing decompositional approaches were developed assuming hidden neuron activation is soft-binary. \emph{We address this limitation in the proposed decompositional approach (NN2Rules) that allows hidden neuron activations to be piecewise linear (which includes ReLU in addition to soft-binary). Our solution exploits a key observation that for neural networks using piecewise linear activations, output of each neuron is a piecewise linear function of the input features.} Our implementation can be found here: \url{https://github.com/groshanlal/NN2Rules}

\section{Related Work}

\paragraph{Decompositional methods} 
Existing decompositional methods for rule extraction from neural networks assume the network consists of a single hidden layer and the hidden neurons use sigmoid activation functions (i.e. their output is soft-binary). When the inputs to the neuron are binary (discretized continuous or categorical inputs), the behaviour of the neuron can be approximated using boolean logic rules. \cite{Fu1994kt} proposed KT (knowledgetron) method that performs a backward pass through the network, starting from the output, searching for boolean rules that confirm or negate the output, proceeding layer by layer through the network to the inputs. \cite{Tsukimoto2000} proposed an algorithm to extract boolean logic rules from a neural network using a forward pass by constructing a truth table for every neuron in terms of its inputs and showed that the computational complexity of their algorithm is polynomial in number of inputs if the rules are restricted to have only a fixed length of attributes. \cite{Lisoa2006osre} extended the approach by \cite{Tsukimoto2000} to categorical attributes by searching the attribute space in orthogonal directions to identify the decision boundaries. \cite{Tsukimoto2001cred} proposed CRED which uses both the learned neural network and the training data to learn decision trees for every neuron using inputs from previous layer. CRED can handle both continuous and discrete inputs. Rule extraction algorithms often use some pruning heuristics on neurons to speed up their rule extraction. \cite{Setiono2000fernn} proposed FERNN to regularize the neural network training to have small weights for insignificant links for pruning the network and speeding up the rule extraction. \cite{Difaconn2010} proposed DIFACONN-miner that combines neural network training with rule extraction using differential evolution algorithm for training and touring ant colony optimization algorithm for rule extraction.

The key difference between NN2Rules and the above approaches is that NN2Rules does not assume that hidden neuron behavior is soft-binary. Instead, it allows hidden neuron behavior to be soft-binary (eg. sigmoid activation) or rectified linear (ReLU). 

\paragraph{Pedagogical methods} 
Pedagogical methods treat the neural network as a black box and sample training points from the neural network to learn a transparent model on it. \cite{Shavlik1995trepan} proposed TREPAN to learn a decision tree by querying training points from the neural network to identify best splits and enrich training data point with less frequently occurring feature attributes. \cite{Sethi2012kdruleex} proposed KDRuleEx, a non-recursive algorithm to learn a decision table from a neural network by sampling data points.
\cite{Saad2007hypinv} proposed HypInv algorithm, which projects a training data point to the decision boundary by following gradients to the level surface of decision boundary. Using the projected data point and the gradient, the decision surface can be approximated by hyper planes, which in turn can provide linear rules for the decisions made by the neural network.
\cite{Thrun95via} proposed Validity Interval Analysis, which starts with a set of interval conditions satisfied by input neurons and output neuron. VIA propagates the interval conditions of input to the max and min of neuron output using linear programming, iteratively on the network in a forward pass. Similarly, given the interval conditions on the output, max and min of input can be found using linear programming, iteratively on the network in a backward pass. Whenever the solution space of the linear program reduces to null, VIA concludes that the rule being verified is false. 
For a neural network with a single hidden layer, \cite{Augasta2012rxren} proposed RxREN to prune for significant hidden neurons. The significant hidden neurons are the ones whose removal would lead to significant errors. RxREN forms rules by inspecting the intervals of significant hidden neurons, whenever they misclassify the data points. 

\paragraph{Eclectic methods}
Eclectic methods are a mixture of both decompositional and pedagogical approaches. They break the network into some coarse parts (larger than neuron level) and extract rules for constituent parts and then put them together to extract rules for the network \cite{Setiono1997rx, Setiono1997neurorule, Hruschka2006, Setiono2008rerx}. All these approaches also assumed soft-binary activation.

\paragraph{Deep Neural Networks}
\cite{Zilke2015} discusses the challenges of extracting rules from deep neural networks. The authors proposed DeepRED, a modified version of CRED for extracting rules from a deep network. The authors also presented modifed versions of FERNN and RxREN for pruning the network for faster rule extraction. Note that their approach assumes soft-binary activation for hidden neurons.

\cite{Hinton2017softtree} presented soft decision trees, an alternate decision tree model which can be trained using back-propagation similar to neural networks. The authors trained soft decision trees by distilling knowledge \cite{Hinton2017distil} from a neural network.

\cite{schaaf2019reg} showed that L1-orthogonal regularization helps in training smaller trees. \cite{Doshi2018treereg, Doshi2020treereg} proposed tree-regularization for training neural networks, which can be approximated with small decision trees. These techniques impose regularization on neural network during training and can be applied to all approaches including our NN2Rules to reduce number of rules.

\section{Problem}

\subsection{Definition}
\textbf{Input:} A neural network model that has been trained for binary classification. \\
\textbf{Output:} A rule set defined on the input features and an associated binary class for each rule.\\
\textbf{Goal:} Find a rule set such that for any test instance, the class predicted by the neural network matches the class predicted by the corresponding rule in the rule set. For better comprehensibility, it is desirable for the rule set to be comprised of a small number of rules, where each rule uses a small number of features.

\subsection{Illustration}
Consider a neural network that predicts if a tomato is ripe or not. It uses $2$ categorical features: Color and Size. Color can be red or yellow, size can be small, medium or big. Let us encode the input data  $[x_1, x_2, y_1, y_2, y_3] =[\text{color-red}, \text{color-yellow}, \text{size-small}, \text{size-medium}, \text{size-big}]$ using one hot encoding. Here is the neural network model that was learned:
$$n_1 = ReLU(4x_1 +  x_2 +  y_1 + 2y_2  + 2y_3 - 5)$$
$$n_2 = ReLU(3x_1 + 2x_2 + 2y_1 + 3y_2 + 3y_3 - 5)$$
$$n_3 = \sigma(n_1 + n_2 - 1)$$

To convert the above neural network to a rule set we begin by first interpreting each hidden neuron (i.e. $n_1$ and $n_2$).

Since color has to be either red or yellow, and size has to be either small, medium or big: $x_1 + x_2 = 1$, and $y_1 + y_2 + y_3 = 1$. This implies
\begin{align*}
n_1 
&= ReLU(4x_1 +  x_2 +  y_1 + 2y_2 + 2y_3 - 5) \\
&= ReLU(3x_1 + y_2 + y_3 - 3) \\
\end{align*}

Since each feature-value pair can be either 0 or 1, i.e., $x_1 \in \{0,1\}$, and $y_2 \in \{0,1\}$, we get
\begin{align*}
n_1 
&= ReLU(3x_1 + y_2 + y_3 - 3) \\ 
& = \begin{cases} 
      y_2 + y_3 & color = red \\
      0   & otherwise 
\end{cases}
\end{align*}

Similarly, 
\begin{align*}
n_2 
&= ReLU(3x_1 + 2x_2 + 2y_1 + 3y_2 + 3y_3 - 5) \\
&= ReLU(x_1 + y_2 + y_3 - 1) \\
& = \begin{cases} 
      x_1 & size = medium \\
      x_1 & size = big \\
      y_2 + y_3 & color = red \\
      0   & otherwise 
\end{cases}
\end{align*}

Notice that $n_1$ and $n_2$ are piecewise linear functions of the input. NN2Rules will leverage this piecewise linear property.

Finally, for the output neuron $n_3 = \sigma(n_1 + n_2 - 1) \geq 0.5$, whenever $n_1 + n_2 - 1 \geq 0$.
\begin{align*}
&n_1 + n_2 - 1 = \\
& \begin{cases} 
      y_2 + y_3 + x_1 - 1 = 1  & color = red \wedge size = big\\
      y_2 + y_3 + x_1 - 1 = 1  & color = red \wedge size = medium\\
      y_2 + y_3 + y_2 + y_3 - 1 = -1 & color = red \wedge size = small\\
      0 + 0 -1 = -1 & otherwise\\
\end{cases}
\end{align*}

Thus, the neural network output for this illustrative example can be represented as the following rule set:
\begin{align*}
n_3 \geq 0.5 = 
& \begin{cases} 
1  & color = red \wedge size = big\\
1  & color = red \wedge size = medium\\
0  & otherwise
\end{cases}
\end{align*}

\section{Our Approach}

In this section we first mention the data and model assumptions made by the NN2Rules, followed by an overview of the method. Finally, we discuss \textsc{LinRule} and \textsc{NeuronRule} algorithms which are the key building blocks of NN2Rules.

\subsection{Data and Model Assumptions}

Decompositional methods for rule extraction make some common assumptions about data and the underlying model. Past approaches \cite{Fu1994kt, Tsukimoto2000, Lisoa2006osre} assume that the data consists of categorical or discretized continuous features and the underlying neural network is shallow with fully connected layers of neurons with sigmoid activation functions. Further, some approaches also approximate the sigmoid activation with hard-binary activation functions. We make the following assumptions on the input data and the network structure:

\textbf{Data:}
    \begin{enumerate}
        \item The input data consists of only categorical features. Numerical features can always be converted into categorical features by binning them appropriately using range-buckets or hash-buckets.
        \item Every categorical feature in the input data can take exactly one discrete value.
    \end{enumerate}
    
Specifically, the input data consists of $m$ categorical features $(f_1, f_2, \ldots, f_m)$. Each of the categorical feature $f_i$s can take exactly one of the $n_i$ different values from the set $V_i = \{v_{i1}, v_{i2}, v_{i3}, \ldots, v_{i{n_i}}\}$.  Each of the input features is represented by a $n_i$ length feature vector using one hot encoding. The $m$ feature vectors are concatenated to form a $n = \sum_{i = 1}^{m} n_i$ length vector $\mathbf{x}^1$ as input to the neural network. 

\textbf{Neural Network Model:}
    \begin{enumerate}
        \item All hidden layer neurons use either sigmoid or ReLU activation.
        \item The output layer neuron uses sigmoid activation.
    \end{enumerate}
    
\begin{align*}
\mathbf{x}^1 &:= \textnormal{Input to the Network} \\
\mathbf{x}^{L+1} &:= \textnormal{Network Prediction}
\end{align*}
For $k = 1, 2, 3 , \ldots, L-1$
\begin{align*}
y^k_i &= \sum_{j} w^k_{ij} x^k_j + b^k_i \\
x^{k+1}_i &= ReLU(y^k_i)
\end{align*}
For the last (output) layer,
\begin{align*}
y^L_i &= \sum_{j} w^L_{ij} x^L_j + b^L_i \\
x^{L+1}_i &= \sigma(y^L_i)
\end{align*}

\subsection{NN2Rules Method}

The key property we leverage in NN2Rules is that if the activation functions are either piecewise linear (eg ReLU) or piecewise constant (soft-binary approximations), then there exists a partitioning of the input features such that the neuron behavior is a linear function of input features in each partition. In other words, \emph{each neuron is a piecewise linear function of the input features}. Our observation is a generalization from the existing decompositional approaches that use the property that if activation functions are piecewise constant (soft-binary) then each neuron can be expressed as a piecewise constant (soft-binary) function of the input features.

Hence, NN2Rules converts every neuron $x_i^k$ (for $k = 2,3,\ldots,L+1$) into a piece-wise linear function of input $\mathbf{x}^1$, where each piece is described by a rule satisfied by a subset of the categorical features $(f_1,f_2,f_3,\ldots,f_m)$. We call the individual linear functions as the neuron behavior corresponding to the rules. Hence, our goal is to convert a neuron into an equivalent list of (rule, behavior) tuples. For example, in the illustration we converted neuron $n_2$ into a list of (rule, behavior) tuples, where one of the tuples was ($color = red$, $y_2 + y_3$).

For the last (output) layer, we would like to know when is $x_i^{L+1} = \sigma(y_i^L) > 0.5$, which is equivalent to asking when is  $y_i^L > 0$. Hence, in the last layer, we can effectively replace the sigmoid activation with ReLU activation and the question of when is the sigmoid function active (greater than $0.5$) is equivalent to the question of when is the ReLU function active (greater than $0$).

NN2Rules approach uses two algorithms as its building blocks:
\begin{enumerate}
    \item \textbf{\textsc{LinRule}}: Given a ReLU activated neuron with $\mathbf{x}^1$ as input, LinRule finds the list of (rule, behavior) tuples for which the neuron is active(or inactive).  
    
    \item \textbf{\textsc{NeuronRule}}: Given a ReLU activated neuron with hidden neurons as input, each of which have their own list of (rule, behavior) tuples, NeuronRule finds the list of (rule, behavior) tuples for which the neuron is active(or inactive). 
    
\end{enumerate}
NN2Rule approach runs \textsc{LinRule} on the first layer of neurons to generate a list of (rule, behavior) tuples. Then, \textsc{NeuronRule} is run on all the hidden layer neurons in a forward pass through the network to get the list of (rule, behavior) tuples for the output neuron.

\subsection{\textsc{LinRule}}

LinRule is used to get the list of (rule, behavior) tuples for the first layer of neurons (i.e. $x^2_i$  $\forall$ $i$ ). The neuron behavior, when it is active, is given by $y^1_i$ (i.e. the pre-activation linear function of the neuron), and $0$ when not active. We obtain the list of rules for neuron activation, using the following two steps (of Algorithm \textsc{LinRule}). 

\paragraph{Step 1}: Given a linear function $\sum_{i = 1}^{n}w_ix_i + b$, we split the weights into $m$ buckets (corresponding to each feature) with $n_i$ weights in the $i^{th}$ bucket.
\begin{align*}
W_1 &= \{w_1, w_2, \ldots, w_{n_1}  \}     \\
W_2 &= \{ w_{n_1 + 1}, w_{n_1 + 2}, \ldots, w_{n_1 + n_2}  \}   \\
\ldots \\
W_m &= \{w_{n_1+...n_{m-1} + 1}, w_{n_{m-1} + 2}, \ldots, w_{n_1 + n_2 + ...n_{m}}  \}
\end{align*}

\paragraph{Step 2: \textsc{SelectWeights} Problem}: 
Given $m$ buckets (corresponding to the $m$ features), with $n_i$ items  in the $i^{th}$ bucket (each item corresponds to a feature value), with a non-negative weight for each item (the weight corresponds to the weight associated with the input feature in the linear function of the neuron), we would like to pick at most one item (corresponding to one feature value) from each bucket to form an itemset such that the total weight of the itemset is greater than a user-specified threshold. The total weight of the itemset is a lower bound on the linear function (due to the non-negative nature of weights). The goal is to find all possible itemsets (and the rules corresponding to the itemsets) with total weight of the itemset greater than the user-specified threshold ($\tau$). 

This is a general combinatorial problem. To put it in a canonical form, we make the weights in each $W_i$ non-negative and sorted in decreasing order. This is accomplished by sorting the weights in each bucket and then subtracting the lowest weight in each bucket from all the weights in the bucket, and adding the subtracted weights back to the bias term. (This is skipped in the pseudo-code for brevity.)

\begin{algorithm}[ht]
\DontPrintSemicolon
\SetAlgoLined
\SetKwInOut{Input}{Input}\SetKwInOut{Output}{Output}
\Input{Weight Vector $\mathbf{w}$ and bias $b$ acting on one-hot encoded categorical input features.}
\Output{A list of tuples of the form (R.rule, R.weight, R.bias), such that whenever R.rule is active, the ReLU activated neuron behavior is governed by R.weight and R.bias}

\BlankLine
RuleList = Empty Rule List\\
\tcc{Split weights into buckets grouped by feature}
\For{$i \textnormal{ in } 1,2,\ldots,m$}{
    $ W_i = \{ w_{n_1+n_2+..n_{i-1} + 1}, \ldots, w_{n_1 + n_2+ ..n_{i}} \}$ \\
    $ V_i = \{ v_{n_1+n_2+..n_{i-1} + 1}, \ldots, v_{n_1 + n_2+ ..n_{i}} \}$ \\
 }
$\mathbf{W} = (W_1, W_2, \ldots, W_m)$   \\
$\mathbf{V} = (V_1, V_2, \ldots, V_m)$   \\
\tcp{get rules that activate ReLU} 
PosRules = \textsc{SelectWeights}$(\mathbf{W},\mathbf{V},-b)$\\
\For{\textnormal{pr in PosRules}}{
    R.rule, R.weight, R.bias   = pr, $\mathbf{w}$, $b$   \\
    RuleList.append(R)
}  
\tcp{get rules that deactivate ReLU} 
NegRules = \textsc{SelectWeights}$(\mathbf{-W},\mathbf{V},b)$\\
\For{\textnormal{nr in NegRules}}{
    R.rule, R.weight, R.bias   = nr, $\mathbf{0}$, $0$   \\
    RuleList.append(R)
}  
 
\BlankLine
\Return{\textnormal{RuleList}}

\caption{\textsc{LinRule}$(\mathbf{w}, b)$}
\end{algorithm}

In our implementation, we constrain that a weight can be picked from $W_i$ only if there has been some weight picked from each of the buckets $W_j$ for $j < i$. This results in our rules always starting with first assigning a value to $f_1$, followed by $f_2$, $f_3$ and so on. This restriction lends itself a Dynamic Programming solution and further helps us in the conjunction step of \textsc{NeuronRule} (Section \ref{subsection:conjunction}).

Note that for every rule starting with $W_1[i_1]$, say $(W_1[i_1], W_2[i_2], W_3[i_3], \ldots W_k[i_k])$, the following rule starting with $W_1[j_1]$: $(W_1[j_1], W_2[j_2], W_3[j_3], \ldots W_k[j_k])$ with $j_1 < i_1, j_2 < i_2, \ldots, j_k < i_k$ would also exceed $\tau$. We exploit this structure present in the problem. In particular, we start by greedily picking the largest weight($W_1[1]$) from $W_1$ and then recursively call \textsc{SelectWeights} for the rest of the weights ($W_2, W_3, \ldots W_m$) with $\tau = \tau - W_1[1]$. For finding rules starting from any $W_1[i_1]$, we take all the rules starting from $W_1[i_1 - 1]$ (we call them preconditions), swap the leading term to be $W_1[i_1]$ and grow the rule greedily by adding new terms till it exceeds $\tau$.   


\begin{algorithm}[ht]
\DontPrintSemicolon
\SetAlgoLined
\SetKwInOut{Input}{Input}\SetKwInOut{Output}{Output}
\Input{weight buckets $\mathbf{W} = (W_1, W_2, \ldots, W_m)$, \\
feature value buckets $\mathbf{V} = (V_1, \ldots, V_m)$, \\ 
threshold $\tau$, \\
Assumptions: $W_i$s are non-negative and sorted in descending order.}
\Output{List of rules. Each rule is a list of feature values(rule.values) and corresponding weights(rule.weights).}
\BlankLine

\tcc{Check for base cases: }
\If{$\sum_{i=1}^{m}\max(W_i) \leq  \tau$}
{return No rule is valid} 

\BlankLine

\If{$\sum_{i=1}^{m}\min(W_i) \geq \tau$}
{return Any rule is valid} 

\BlankLine

\tcc{Start with rules on first feature and extend till it reaches threshold}
precondition = [Rule($W_1$[1], $V_1$[1])]\\
RL = [] \\

\For{$i = 1,2,3,\ldots,n_1$}{
\tcc{All rules starting with first feature taking ith value}
RLi = []\\
\For{\textnormal{rule in precondition}}{
rule.weights[1] = $W_1[i]$\\
rule.values[1] = $V_1[i]$\\
k = len(rule)\\
$\mathbf{W} = (W_{k+1},W_{k+2},\ldots, W_m )$\\
$\mathbf{V} = (V_{k+1},V_{k+2},\ldots, V_m )$\\
$\tau'$ = sum(rule.weights) \\
ruleExtension = \textsc{SelectWeights}($\mathbf{W},\mathbf{V},\tau - \tau'$))\\	
Grow rule with suffixes from ruleExtension\\
Append all extended rules to RLi
}
RL = RL.append(RLi)\\
preconditions = RLi\\
}
return RL\\
\caption{\textsc{SelectWeights}$(\mathbf{W}, \mathbf{V}, \tau)$}
\end{algorithm}

\subsection{\textsc{NeuronRule}}
\textsc{NeuronRule} is used to get the list of (rule, behavior) tuples for the hidden layer of neurons (i.e. $x^k_i$  $\forall$ $i$ and $\forall$ $k > 2$). For a neuron in the hidden layers, each of its input neuron also has its own list of (rule, behavior) tuples. In a forward pass, NN2Rules combines these rules and behaviors of the input neurons to obtain the list of (rule, behavior) tuples for the current neuron. We accomplish this task with the following 3 steps (of Algorithm \textsc{NeuronRule}):
\begin{enumerate}
    \item \textbf{Combining rules of input neurons}: We perform a conjunction of the rules from the input neurons to obtain the equivalent input conditions (i.e. the rule set) of the current neuron. In general for $p$ rulelists corresponding to the $p$ input neurons, and each rulelist with $n$ rules, there can be $n^p$ combinations of rules for which we need to perform conjunction. In the section 4.5, we show that, we can significantly speed this up, since in our implementation, rules always start with assigning a value to $f_1$, followed by  $f_2$ and so on.
    
    \item \textbf{Combining neuron behavior}: Given the behavior of each input neuron as weights and bias in terms of input features, we perform a linear combination of behavior of input neurons to obtain the behavior of the current neuron. Note that the new behavior is again a linear function of input features for each rule (obtained using conjunction of rules of input neurons).
   
    \item \textbf{Extracting rules of the neuron, given the input conditions}: Given the input condition of the neuron (obtained from combining rules of input neurons) and the corresponding neuron behavior (obtained from combining neuron behaviors), we finally extract a rule list which extends the input condition and activates the neuron behavior. This is performed using \textsc{LinRule}, since the neuron behavior is a linear function on input features (for each partition of input space defined by a single rule).
\end{enumerate}

\begin{algorithm}[ht]
\DontPrintSemicolon
\SetAlgoLined

\SetKwInOut{Input}{Input}\SetKwInOut{Output}{Output}

\Input{Weight Vector $\mathbf{w}$ and bias $b$ acting on input neurons which have their own rules in the form (R.rule, R.weight, R.bias).}

\Output{A list of tuples of the form (R.rule, R.weight, R.bias), such that whenever R.rule is active, the ReLU activated neuron behavior is governed by R.weight and R.bias.}

\BlankLine
RuleList = Empty Rule List\\
p = number of input neurons\\
\For{$(R_1, R_2, \ldots R_p)$ \textnormal{ in } $\prod_{j=1}^{p} \textnormal{InputNeuronRules}_j$}{
    \tcc{Combining rules of input neurons: Logical AND of rules from previous layer neurons.}
    PreCondition = \textsc{AND}$(R_1.rule, R_2.rule, \ldots, R_p.rule)$

    \BlankLine
    \tcc{Combining neuron behavior: Get weights and bias in terms of input features}
    $\mathbf{w}' = \sum_{j=1}^{p} w_{j}  $R$_j$.weights \\ 
    $b' = b + \sum_{j=1}^{p} w_{j}  $R$_j$.bias \\

    \BlankLine
    \tcc{Extract rules of the neuron in current layer}
    NeuronRules = \textsc{LinRule}$(\mathbf{w}',  b')$ \\
    
    \For{\textnormal{R in NeuronRules}}{
        R.rule = \textsc{AND}(PreCondition, R.rule)
    }
    
    \BlankLine
    RuleList.append(NeuronRules)
}

\BlankLine

\Return{\textnormal{RuleList}}

\caption{\textsc{NeuronRule}$(\mathbf{w}, b)$}
\end{algorithm}

\subsection{Conjunction} \label{subsection:conjunction}
When rules follow the order of features, (i.e, rules always start with assigning a value to $f_1$, followed by  $f_2$ and so on), conjunction of two rules is the longer rule of the two if and only if the smaller rule is prefix of the longer rule or is null otherwise. For example, $(color=red, size=small)$ $AND$ $(color=red) = (color=red, size=small)$, while $(color=red, size=small)$ $AND$ $(color=yellow) = null$.

For performing conjunction of two such rulelists with $n$ rules each, we first sort each of the lists according to lexicographic ordering of rule terms. We then use two pointers one from each rule list starting from the lexicographically smallest rule. Whenever the smaller rule of the two is a prefix of the longer rule, we note down the longer rule in the result and advance the pointer corresponding to the longer rule. If the smaller rule is not a prefix of the longer rule, we advance the pointer corresponding to the lexicographically smaller of the two. Hence, we can obtain all the rule conjunctions from the two rule lists by passing once through both the lists. The resulting list of rules would contain at most $2n$ rules.

For performing conjunction of $p$ such rule lists with $n$ rules each, we first sort each of the lists, like before in $O(pn\log{n})$ steps. We take two rule lists at a time and perform conjunction like before in $O(pn)$ steps. Then we are left with $p/2$ rule lists each with atmost $2n$ rules. Performing $k$ such iterations leaves us with $p/2^k$ rule lists each with at most $2^kn$ rules in a total of $O(pnk)$ steps. Hence, in $O(pn\log{p})$ steps, we can perform conjunction of the $p$ rule lists. Thus, it takes a total of $O(pn\log{n} + pn\log{p}) = O(pn\log(pn))$ steps to perform the conjunction. Note that this is much faster than performing conjunction across all $O(n^p)$ rule combinations, which would be needed if the ordering constraint for the rules is not used.

\section{Results}
In this section we discuss the performance of NN2Rules in terms of the fidelity and comprehensibility of the rule list generated by it on 4 benchmark datasets.

\subsection{Datasets}
We evaluate our method, NN2Rules, on 4 binary classification benchmark datasets selected from the UCI Machine Learning repository \cite{UCIRepository}. In each of these datasets, the numerical features were discretized into three bins. More details on number of features and categories of each of these datasets can be found in Table \ref{table:data}.

\begin{itemize}
    \item Adult Income. The prediction task is to determine whether the income exceeds \$50K/yr based on the 1994 census data. We preprocessed the dataset by dropping some features (fnlwgt, education-num, workclass, relationship, and race) and simplifying some sparse categories by combining them together. For example, we simplify the sparse native-countries feature into US and non-US categories.  
    
    \item Contraception: The prediction task is to determine whether women choose to use contraception based on their demographic and socio-economic features from the 1987 National Indonesia Contraceptive Prevalence Survey. 

    \item Nursery: The prediction task is to determine if a student is admitted to a nursery school or not based on parents socio-economic features. 
    
    \item Cars: The prediction task is to determine whether a car is acceptable or not based on its price and technical characteristics.
\end{itemize}

\begin{table}
\begin{tabular}{ |c|c|c|} 
 \hline
 \textbf{Dataset} & \textbf{Number of Features} & \textbf{Number of Categories} \\
    &   & \textbf{(per feature)} \\
 \hline
 Adult Income & 9 &[3, 3, 3, 3, 3, 14, 2, 8, 2] \\  \hline
 Contraception & 9 & [3, 2, 4, 4, 3, 4, 2, 2, 4]  \\  \hline
 Nursery & 8 & [3, 3, 3, 5, 2, 3, 4, 4] \\  \hline
 Cars & 6 & [3, 3, 4, 3, 4, 4] \\  \hline
\end{tabular}
\caption{Description of UCI Benchmark Datasets used in our experiments. In Adult Income dataset, the numerical features are converted into categorical features by binning. Contraception, Nursery and Cars datasets contain only categorical features.}
\label{table:data}
\end{table}

\subsection{Model}
We train a neural network with $2$ hidden layers with $6$ and $3$ neurons in each layer respectively. We run our method, NN2Rules algorithm on this neural network to get a rule list corresponding to the positive class prediction by the neural network. We check the support (i.e., the number of data points satisfying the rule) for each rule on the training data. We sort the rules in decreasing order of support on training data. We call these rules \textbf{NN2Rules(Full)}. The resulting rule list can be long with many rules. We also consider a subset of these rules which have non-zero support on the training data. We call these rules \textbf{NN2Rules(Support)}.

\subsection{Experiments}

We compare the performance of NN2Rules(Full) and NN2Rules(Support) against decision tree models since they are the most commonly used interpretable models. We use two types of tree models: 1) A decision tree trained independently on the training data. 2) A tree trained using labels from the neural network (TREPAN). We use a simple version of TREPAN which only uses the training data used for training the underlying neural network to train the surrogate tree model. Note that, TREPAN is a pedagogical approach which treats the underlying neural network as a black box and uses only the labels generated by the neural network to train the surrogate tree model. We evaluate the rule lists generated by NN2Rules against the decision tree and TREPAN surrogate on three key explainability aspects:
\begin{enumerate}
    \item \textbf{Fidelity}: Fidelity is a measure of the accuracy of the model explainer. We define Fidelity as the fraction of test instances for which the decision from the Rule List obtained using NN2Rules matches with the neural network model prediction. 
    
    \item \textbf{Comprehensibility}: Comprehensibility is a measure of how easy is the model explainer for a human to grasp. We use the number of rules corresponding to positive instances in the rule list as a comprehensibility measure. 
    
    \item \textbf{Understanding model errors}: We use the fidelity on the test instances, for which the model output disagrees with the test label, as a measure of model error understanding. Fidelity on the model errors in the test data is specifically interesting because often model explanations are used for debugging / understanding model errors.
\end{enumerate}

\subsubsection{Fidelity}

\begin{figure}[ht]
	\centering
	\includegraphics[width=0.55\textwidth]{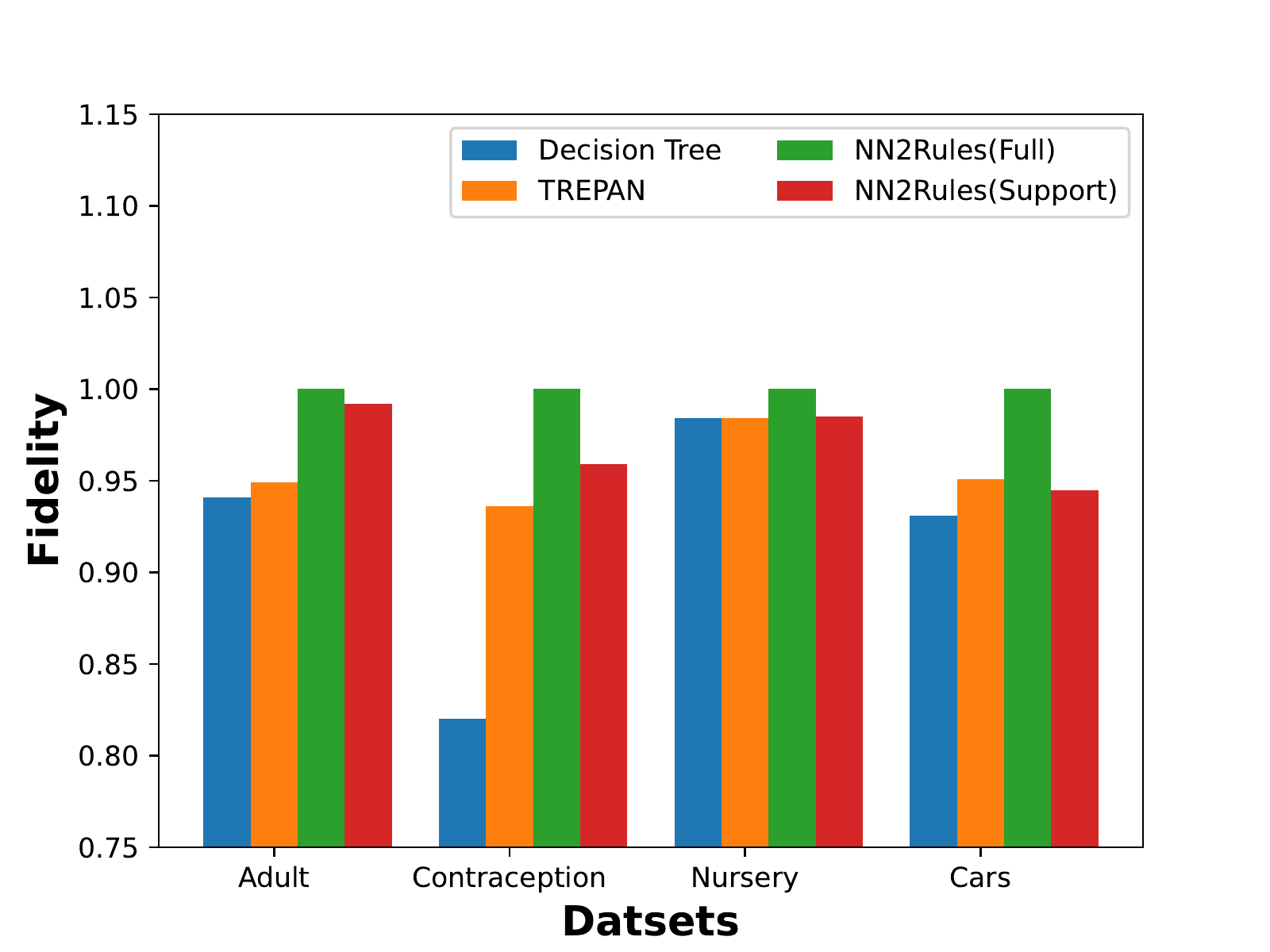}
	
	\vspace{0.5cm}

	\begin{tabular}{ |c|c|c|c|c|}
	 \hline
	 \textbf{Dataset} & \textbf{Decision} & \textbf{TREPAN} & \textbf{NN2Rules} & \textbf{NN2Rules} \\
	    & \textbf{Tree} &   & \textbf{(Full)}  &  \textbf{(Support)} \\
	 \hline
	 Adult Income & 0.941 & 0.949 & 1.000 & 0.992   \\  \hline
	 Contraception & 0.820 & 0.936 & 1.000 & 0.959  \\  \hline
	 Nursery & 0.984 & 0.984 & 1.000 & 0.985\\  \hline
	 Cars & 0.931 & 0.951 & 1.000 & 0.945\\  \hline
	\end{tabular}

	\caption{Comparing fidelity of NN2Rules(Full) and NN2Rules(Support) with Decision Tree and TREPAN. NN2Rules(Full) has full (100\%) fidelity by design.}

	\label{fig:fidelity_bar}
\end{figure}

NN2Rules produces rule lists which always agree with the neural network, since the the rule list generated by NN2Rules is a complete decomposition of the underlying neural network. Thus, the fidelity of NN2Rules(Full) with the entire rule list is 100\%. NN2Rules(Support) uses a subset of rules generated by NN2Rules(Full) and hence has fidelity less than 100\%. Furthermore, in Figure \ref{fig:fidelity_bar}, we observe that for most datasets, NN2Rules(Support) produces rule lists of higher fidelity compared to Decision Tree and TREPAN. 
 
\begin{figure}[ht]
	\centering
	\includegraphics[width=0.55\textwidth]{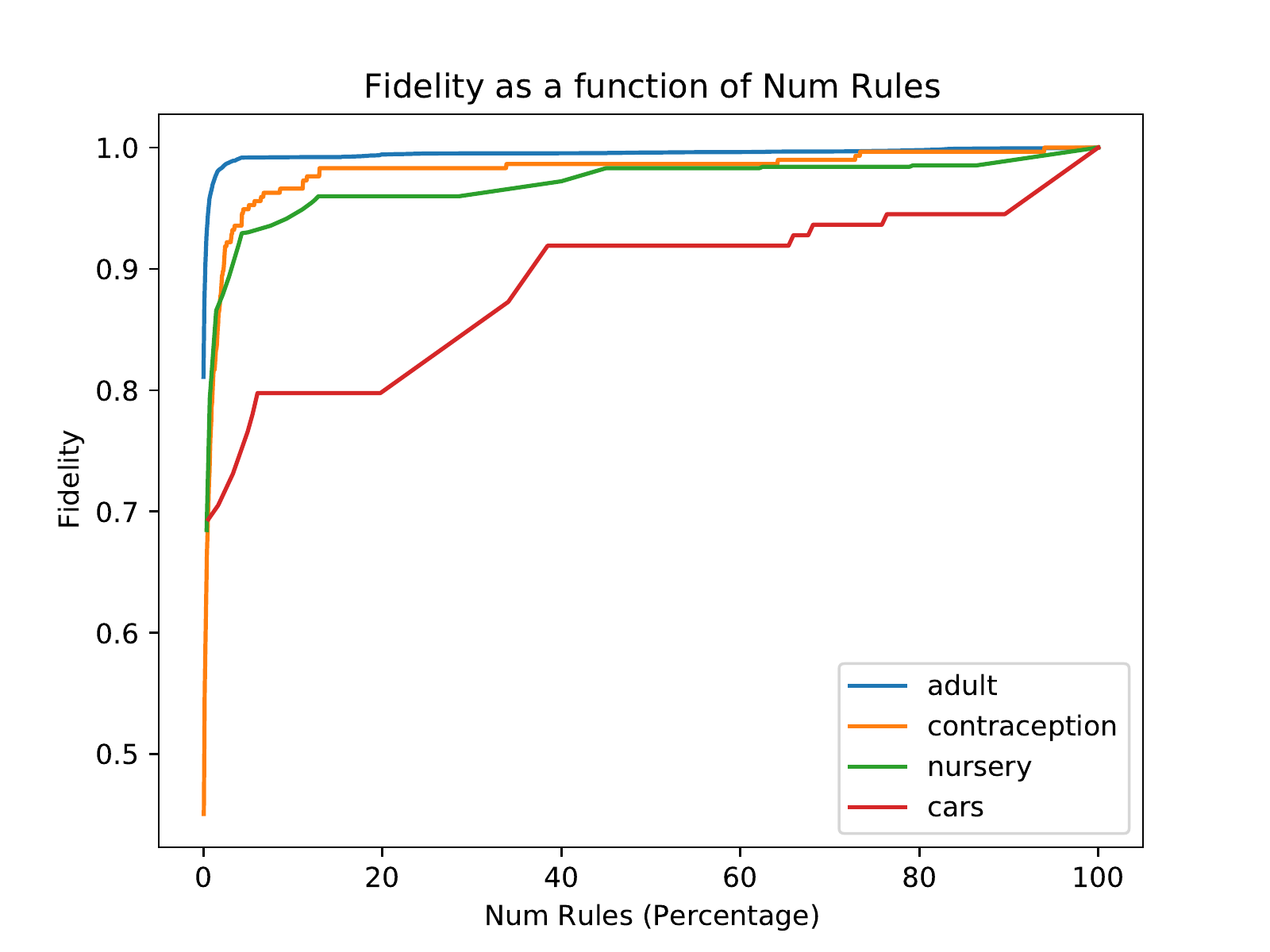}
	\caption{Tradeoff of fidelity vs comprehensibility with number of rules generated by NN2Rules method. X axis shows number of rules generated by NN2Rules, with 0\% indicating all labels scored as negative and 100\% indicating NN2Rules(Full). We can note that the first few rules capture most of the fidelity. Steeper curve is better, i.e with (say) 20\% rules from the respective models, adult income has the highest fidelity explanations followed by contraception, nursery and cars.}
	\label{fig:fidelity}
\end{figure}

\subsubsection{Accuracy}

Since NN2Rules(Full) produces rule lists of 100\% fidelity, the accuracy of NN2Rules(Full) on test data is the same as that of the underlying neural network. Using a subset of rules as in NN2Rules(Support) results in a slightly lower accuracy than the underlying network with the benefit of being more comprehensible (easier to understand for a human). We also observe from Table \ref{table:accuracy} that the accuracy of the underlying neural network on test data is higher than the other interpretable models (like decision tree, TREPAN) for most datasets. 

\begin{table}
	\begin{tabular}{ |c|c|c|c|c|c|} 
		\hline
		\textbf{Dataset} & \textbf{Decision} & \textbf{TREPAN} & \textbf{Neural} & \textbf{NN2Rules} & \textbf{NN2Rules} \\
		& \textbf{Tree} &   & \textbf{Network} & \textbf{(Full)}  &  \textbf{(Support)} \\
		\hline
		Adult Income & 0.843 & 0.845 & 0.854 & 0.854 & 0.851 \\  \hline
		Contraception & 0.610 & 0.651 & 0.647 & 0.647 & 0.654 \\  \hline
		Nursery & 0.975 & 0.975 & 0.988 & 0.988 & 0.985 \\  \hline
		Cars & 0.934 & 0.960 & 0.962 & 0.962 & 0.942 \\  \hline
	\end{tabular}
	\caption{Comparing accuracy of NN2Rules(Full) and NN2Rules(Support) with the underlying Neural Network, Decision Tree and TREPAN. NN2Rules(Full) has the same accuracy as the underlying Neural Network by design.}
	\label{table:accuracy}
\end{table}

\begin{figure}[ht]
    \centering
    \includegraphics[width=0.55\textwidth]{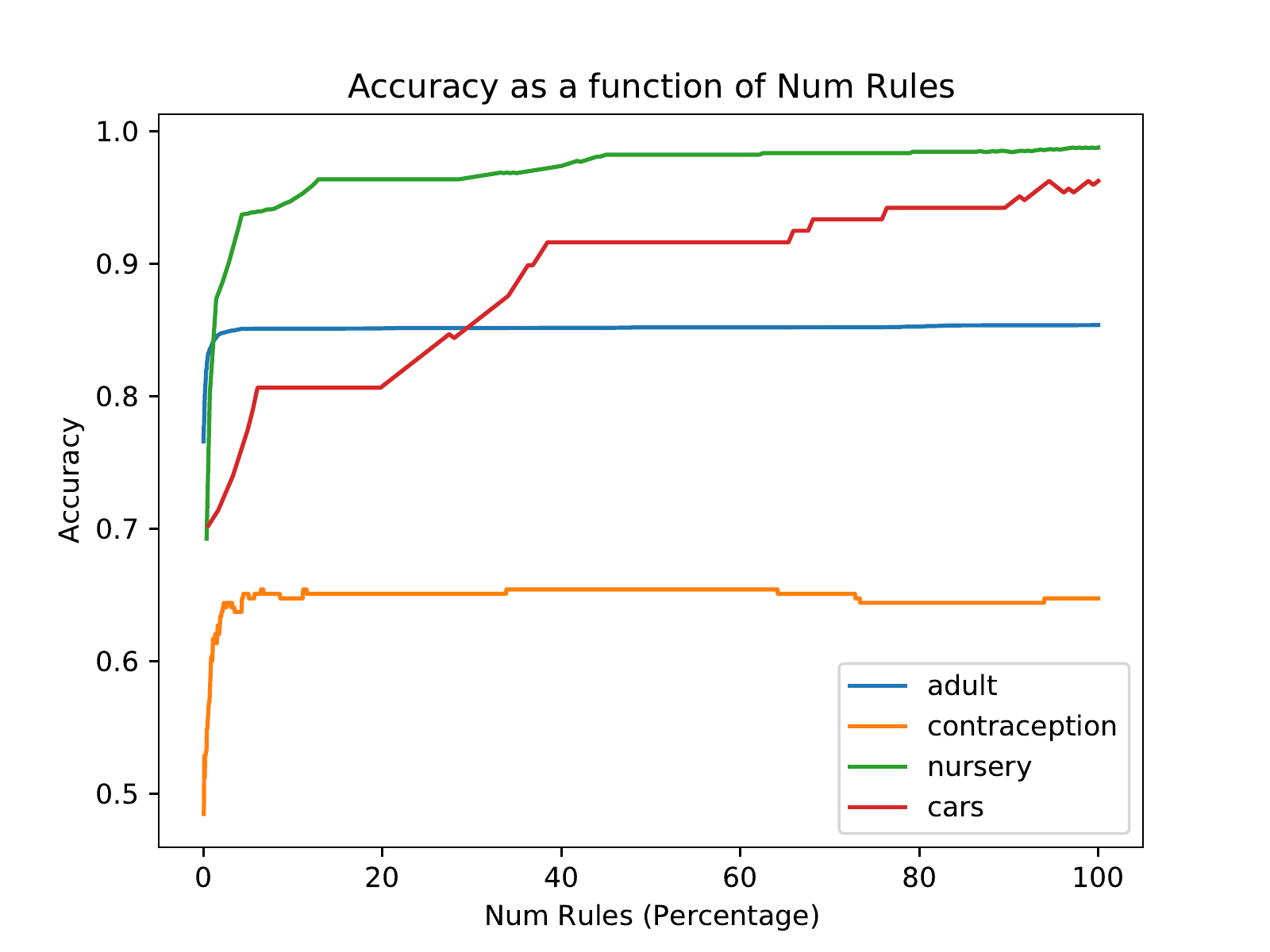}
    \caption{Tradeoff of accuracy vs comprehensibility with number of rules generated by NN2Rules method. X axis shows number of rules generated by NN2Rules, with 0\% indicating all labels scored as negative and 100\% indicating NN2Rules(Full). We can note that the first few rules capture most of the accuracy (steeper curve is better). At 100\% rules, accuracy of the rule list is same as that of the underlying neural network trained on respective datasets.}
    \label{fig:accuracy}
\end{figure}

\subsubsection{Comprehensibility}

In general, NN2Rules generates a long list of rules for which the neural network gives a positive label. However, the rules generated is still only a small subset of the total possible rules. Since all the input data sets are categorical in nature, we can compute a theoretical limit on the number of rules as the product of number of categories for each feature.  We call this Num Rules Max in Table \ref{table:numrules}. 

Since it is often enough to generates rules for positive instances, another baseline metric for comprehensibility is the number of unique positive instances in training data. The feature-category combination corresponding to each positive instance can be memorized as a rule. We call this Num Rules Memorization in Table \ref{table:numrules}.

In Table \ref{table:numrules}, we can observe that the number of rules in NN2Rules(Full) is only a small fraction of Num Rules Max and the number of rules in NN2Rules(Support) is only a small fraction of number of rules in NN2Rules(Full). Furthermore, in Figure \ref{fig:fidelity} and Figure \ref{fig:accuracy}, we can note that the first few rules of NN2Rules(Full), i.e, NN2Rules(Support) capture most of the information of the model with high fidelity and accuracy. 
Figure \ref{fig:fidelity} shows the trade-off between fidelity and comprehensibility(number of rules) for the rules generated by NN2Rules for different datasets. 
Figure \ref{fig:accuracy} shows the trade-off between accuracy and comprehensibility(number of rules) for the rules generated by NN2Rules for different datasets. 

\begin{table*}
\begin{tabular}{ |c|c|c|c|c|c|} 
 \hline
 \textbf{Dataset} &  \textbf{Num Rules} & \textbf{Num Rules}& \textbf{NN2Rules} & \textbf{NN2Rules} \\ 
 & \textbf{Max} & \textbf{ Memorization} & \textbf{(Full)} & \textbf{(Support)} \\
 \hline
 Adult Income & 108864 & 1793 & 12802 & 553 \\  \hline
 Contraception & 18432 & 344 & 2064 & 134 \\  \hline
 Nursery & 12960 & 6663 & 280 & 242\\  \hline
 Cars & 1728 & 414 & 182 & 163\\  \hline
\end{tabular}
\caption{Comparing the number of rules (comprehensibility) of NN2Rules(Full) and NN2Rules(Support) with the maximum number of rules that can be constructed from the datasets and the number of rules that can be formed by memorizing the training data. We can observe that NN2Rules(Full) and NN2Rules(Support) only learn a small subset of these rules.}
\label{table:numrules}
\end{table*}

\subsubsection{Explaining Errors}

One of the important reasons to develop model explainers is to be able to explore model behavior for instances where the model makes an error with respect to ground truth labels for debugging. An advantage of having an explainer model with 100\% fidelity is that it can be used to investigate errors made by the underlying neural network. 

\begin{figure}[ht]
    \centering
    \includegraphics[width=0.55\textwidth]{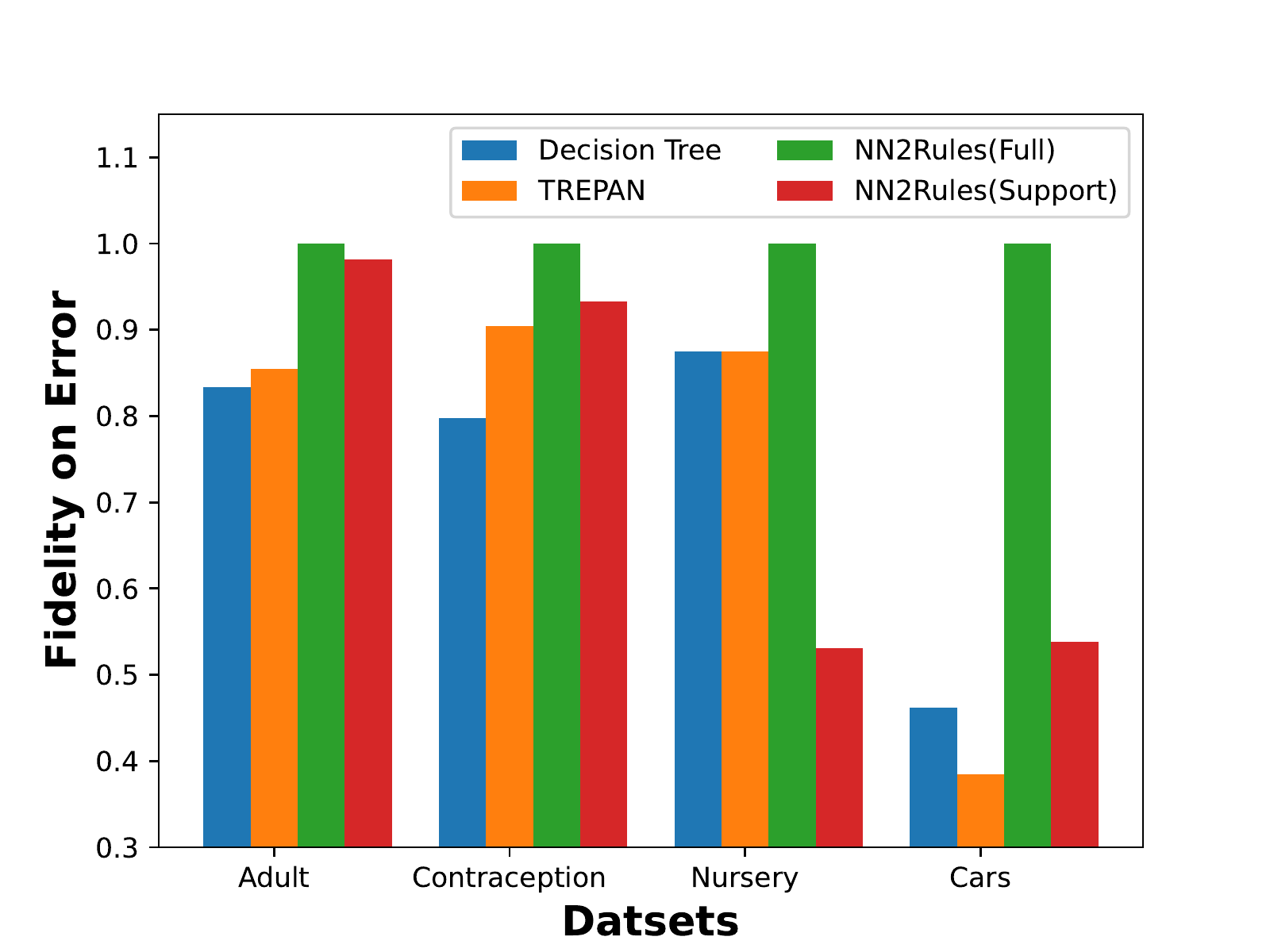}
    	
    \vspace{0.5cm}
    
	\begin{tabular}{ |c|c|c|c|c|} 
	\hline
	\textbf{Dataset} & \textbf{Decision} & \textbf{TREPAN} & \textbf{NN2Rules} & \textbf{NN2Rules} \\
	& \textbf{Tree} &   & \textbf{(Full)}  &  \textbf{(Support)} \\
	\hline
	Adult Income & 0.834 & 0.855 & 1.000 & 0.982   \\  \hline
	Contraception & 0.798 & 0.904 & 1.000 & 0.933  \\  \hline
	Nursery & 0.875 & 0.875 & 1.000 & 0.531  \\  \hline
	Cars & 0.462 & 0.385 & 1.000 & 0.538  \\  \hline
	\end{tabular}

	\caption{Comparing fidelity of NN2Rules(Full) and NN2Rules(Support) with Decision Tree and TREPAN on the test errors. NN2Rules(Full) has full (100\%) fidelity by design. NN2Rules(Support) is able to explain the errors better in datasets where the fidelity-comprehensibility tradeoff is steeper in Figure \ref{fig:fidelity}.}

    \label{fig:fidelity_error_bar}
\end{figure}

Pedagogical methods treat the underlying model as a black box and use the training data and labels estimated by the underlying model to build model explainers. The errors made by the underlying model on the test data come from falsely learnt patterns or patterns which cannot be easily learnt from the training data, due to lack of enough data samples, or noisy data samples. Pedagogical model explainers, which also rely on training data, are not best suited for explaining model errors on test data. In contrast, Decompositional methods like NN2Rules solely rely on the underlying model parameters and do not use data distribution at all. This intuition is also supported in Figure \ref{fig:fidelity_error_bar}, where we observe that the gap in fidelity of other approaches with respect to NN2Rules(Full) is higher for the error test data compared to the full test data in Figure \ref{fig:fidelity_bar}.

NN2Rules(Support) uses training data to select a smaller subset of rules and therefore has lesser fidelity on test errors. We observe that its ability to explain model errors is correlated to how steep the fidelity curve is in Figure \ref{fig:fidelity}. For datasets where the fidelity curve is steeper, the first few rules from NN2Rules(Full) capture most of the fidelity and therefore NN2Rules(Support) is able to better explain the model errors than other baselines like pedagogical approaches (TREPAN) and decision tree.

\section{Conclusion and Future Work}

In this paper we presented NN2Rules, a new decompositional approach to convert a neural network model to a rule list. NN2Rules decomposes the behavior of each hidden neuron as a collection of linear function of the inputs for partitions of the input feature-space. In contrast, prior approaches decomposed the behavior of hidden neurons as either 1 or 0 for partitions of feature-space because of which they only worked for neurons with soft-binary (sigmoid) activation functions. The ability of NN2Rules to track neuron behavior as linear function of inputs generalizes decompositional approaches to work with piecewise linear activation functions (e.g. the commonly used ReLU). Our experiments show that NN2Rules achieves 100\% fidelity as expected and is better at explaining model behavior for erroneous test instances than pedagogical approaches. Moreover, our results indicate that while NN2Rules may generate a larger rule list, often only the top few rules (sorted by their coverage) are instrumental in explaining the test data. Hence, the rule lists generated from NN2Rules can be safely pruned in the interest of achieving better comprehensibility.

NN2Rules is a step towards improving neural network interpretability. It uses a fundamental building block- \textsc{SelectWeights}. Our current solution to \textsc{SelectWeights} is correct (i.e. all discovered rules are valid) and complete (i.e. all valid rules are discovered). Future research can improve \textsc{SelectWeights} in two aspects (i) scalability, and (ii) comprehensibility. For scalability one can either (i) leverage the intrinsic parallelism as the algorithm is independently applied for each neuron of a given layer, or (ii) modify the algorithm itself to improve its time complexity. Our current solution solves a constrained version of the original \textsc{SelectWeights} problem, i.e., we pick features in pre-specified order. While this constraint helped leverage an optimal subproblem structure in recursion as well as optimized Conjunction (AND) operation, it also undesirably leads to fragmentation of rules (i.e. a rule is broken into multiple rules of higher rule-width and smaller coverage). Future research is needed to find ways to either fix the ordering of features to minimize fragmentation or explore solutions to (the unconstrained version of) the \textsc{SelectWeights}.





\bibliographystyle{plainnat}
\bibliography{main}

\end{document}